\documentclass[conference]{IEEEtran}
\IEEEoverridecommandlockouts

\usepackage{cite}
\usepackage{amsmath,amssymb,amsfonts}
\usepackage{graphicx}
\usepackage{textcomp}
\usepackage{xcolor}
\usepackage{booktabs}
\usepackage{multirow}
\usepackage{url}
\usepackage{array}
\usepackage{enumitem}
\usepackage{subcaption}
\usepackage{xurl}
\usepackage{placeins}

\def\BibTeX{{\rm B\kern-.05em{\sc i\kern-.025em b}\kern-.08em
    T\kern-.1667em\lower.7ex\hbox{E}\kern-.125emX}}

\graphicspath{{figures/}}
\newcolumntype{Y}{>{\raggedright\arraybackslash\hspace{0pt}}p{0.34\columnwidth}}

\begin{document}

\title{Balancing Chain-of-Thought and Answer Supervision for Small Language Models on Medical Question Answering}

\author{\IEEEauthorblockN{Le Zhou}
\IEEEauthorblockA{\textit{Pace University} \\
New York City, USA \\
lz25261n@pace.edu}
\and
\IEEEauthorblockN{Xiao Luo}
\IEEEauthorblockA{
\textit{Southern Methodist University}\\
Dallas, USA \\
xnluo@smu.edu}
\and
\IEEEauthorblockN{Chinmay Chandra}
\IEEEauthorblockA{
\textit{Pace University}\\
New York City, USA \\
cc55048n@pace.edu}
\and
\IEEEauthorblockN{Zhan Zhang}
\IEEEauthorblockA{
\textit{Pace University}\\
New York City, USA \\
zzhang@pace.edu}
}

\maketitle

\begin{abstract}
Small language models ($\leq$9B parameters) can support local medical question answering (QA) while reducing privacy and latency concerns, but effective fine-tuning remains challenging. This paper compares two chain-of-thought (CoT) supervised fine-tuning methods: Uniform-across-Sequence (UaS), which applies equal loss weighting to the full CoT-and-answer sequence, and Segment-weighted (SwS), which separately weights CoT and answer losses. We evaluate six general- and medical-domain models on multiple-choice QA (MedQA and EMS-MCQA) and free-text QA (Med-O1). Results show that neither method is universally best: on multiple-choice QA (MedQA and EMS-MCQA), SwS generally performs better, with general-purpose models benefiting most; notably, Qwen3.5-4B improves by up to 44.8\% over its zero-shot baseline on MedQA. In contrast, medical-domain models show smaller or even negative gains on MCQs. On free-text QA (Med-O1), UaS performs best for nearly all models and provides consistent improvements across both model types, with the 3B Llama-3.2 model approaching the performance of much larger 7B medical-domain models. We also find that excessively long CoT generation is associated with lower performance in some models on either multiple-choice or free-text QA tasks, suggesting that limiting the CoT length during training may benefit selected models. Overall, the results show that CoT loss weighting should be matched to the task format and the knowledge already encoded in the base model, rather than using a single fine-tuning strategy for all clinical QA tasks.
\end{abstract}


\section{Introduction}

Large language models (LLMs) have shown substantial promise for medical question-answering (QA), supporting applications ranging from clinical decision support to medical education and preparation for medical licensing examinations  \cite{singhal2023large,nori2023capabilities,singhal2025toward}. However, deploying these capabilities in real-world clinical and educational settings increasingly demands small language models (SLMs) that can be deployed locally rather than relying on large proprietary systems, owing to constraints around computational cost, latency, data privacy, and local hosting requirements in healthcare environments \cite{singhal2025toward,ren2026medical}. Given the benefits of SLMs, recent research has increasingly explored effective approaches for training and fine-tuning these models for medical QA tasks \cite{kim2025small,ferrazzi2026small}. Conventional instruction tuning trains SLMs to map a clinical question directly to a final answer, optimizing for benchmark accuracy without requiring the model to explicitly articulate the reasoning underlying its prediction \cite{wei2022flan,Chung2024ScalingIF}. Although this approach can improve task performance, training only on final answers provides little visibility into whether the model's reasoning process is clinically sound and offers no mechanism for identifying flawed reasoning that may nevertheless produce a correct answer by chance \cite{cobbe2021training}. Chain-of-thought (CoT) supervision, in which models are trained to generate an explicit reasoning trace alongside the final answer, has emerged as a promising alternative. Foundational studies demonstrated that CoT supervision can substantially improve performance on complex reasoning tasks \cite{wei2022chain,kojima2022large}, while more recent work has shown that supervising intermediate reasoning steps and process traces can improve reasoning quality, robustness, and interpretability compared with training based only on final answers \cite{lightman2023lets,turpin2023language}. As a result, supervised fine-tuning (SFT) that incorporates CoT reasoning has increasingly been used to improve the reasoning capabilities of language models. Despite this shift, an important training design question remains largely unexplored: when optimizing models to generate both reasoning traces and final answers, how should the loss associated with these two components be balanced? This question is particularly important for SLMs, whose limited representational capacity may require trade-offs between learning effective reasoning processes and producing accurate final answers.

In this work, we address this question by implementing different loss weighting strategies that jointly fine-tune language models on CoT rationales and final answers, allowing the relative contribution of reasoning versus answer tokens to be explicitly controlled during training. We apply this approach across two broad categories of SLMs: three medical-domain models (BioMistral-7B, Meerkat-8B, and Lingshu-7B) and three general-purpose models with no medical pretraining (Llama-3.2-3B, Qwen3.5-4B, and Qwen3.5-9B). This selection allows us to examine not only whether the joint CoT-and-answer fine-tuning improves performance within a given model category, but also whether its benefits depend on the model's prior medical domain exposure. We evaluate our approach on three datasets spanning different clinical reasoning demands: MedQA, a widely used USMLE-style multiple-choice benchmark; Med-O1, which pairs questions with longer, more complex CoT traces and free-form responses; and EMS-MCQA, a multiple-choice benchmark grounded in emergency medical services protocols.

Our results show that joint CoT-and-answer fine-tuning improves medical QA performance, although the magnitude of improvement varies across models and tasks. Notably, general-purpose SLMs (Llama-3.2-3B, Qwen3.5-4B, and Qwen3.5-9B) generally show larger gains than medical-domain models (BioMistral-7B, Lingshu-7B, and Meerkat-8B), suggesting that explicit reasoning supervision may help compensate, at least in part, for the absence of medical domain pretraining. However, this benefit is not universal, as models with stronger baseline medical performance show smaller or even negative gains in some settings. These findings demonstrate that the way reasoning and answer losses are balanced during fine-tuning can influence model performance and should therefore be considered in relation to both the underlying model and the medical question answering task.

In summary, this paper makes the following contributions:
\begin{itemize}
    \item We examine different loss weighting strategies for jointly fine-tuning SLMs on CoT rationales and final answers, enabling explicit control over the relative contribution of reasoning and answer supervision during training.
    \item  We systematically evaluate these strategies across six SLMs, including both medical-domain and general-purpose models ranging from 3B to 9B parameters, on three medical QA datasets.
    \item  We find that excessively long CoT generation is associated with lower accuracy in some model–task settings, suggesting that adaptive training strategies based on the CoT length may benefit certain models.
\end{itemize}

In summary, our work offers practical guidance on how to balance reasoning and answer supervision when adapting compact, locally deployable SLMs for medical QA, particularly in settings where both answer accuracy and the quality of clinical reasoning are critical.

\section{Related Work}

Research on fine-tuning SLMs for medical QA has increasingly focused on developing compact models that generate clinically grounded answers while remaining computationally efficient for deployment. Existing work can be broadly categorized into answer-only instruction tuning and, more recently, rationale-supervised CoT fine-tuning.

Early work primarily relied on answer-only instruction tuning, where models are trained to generate correct answers without explicit supervision of their intermediate reasoning processes. Representative examples include MedAlpaca, which extends Stanford Alpaca by fine-tuning LLaMA-7B/13B on curated medical QA pairs, flashcards, and dialogue datasets to produce open-source medical QA and chatbot models \cite{han2023medalpaca}. Similarly, ChatDoctor fine-tunes LLaMA on medical instruction datasets to develop a conversational medical assistant \cite{li2023chatdoctor}. PMC-LLaMA further improves domain adaptation by continually pretraining LLaMA on biomedical literature from PubMed Central before instruction tuning, resulting in improved performance on medical QA and biomedical knowledge retrieval \cite{wu2023pmcllama}. Likewise, BioMistral applies continual pretraining to the Mistral-7B architecture using biomedical corpora to develop an open-source biomedical language model with strong performance across multiple medical benchmarks \cite{labrak2024biomistral}. Comparative evaluations of models such as PMC-LLaMA, BioMistral, LLaMA-2, and Mistral under both zero-shot and LoRA fine-tuning settings have shown that instruction tuning substantially improves answer accuracy. However, these approaches provide limited evidence regarding whether the models have learned clinically meaningful reasoning or are simply producing correct answers \cite{garikipati2024openmedlm}.

To address this gap, recent studies have begun fine-tuning models directly on CoT rationales rather than answer labels alone. Unlike answer-only instruction tuning, rationale-supervised fine-tuning explicitly trains models to generate intermediate reasoning steps together with the final answer, encouraging the model to learn the clinical reasoning process underlying its predictions. For example, the Meerkat models were fine-tuned using GPT-4-generated CoT rationales for USMLE-style medical questions and achieved substantial improvements over answer-only baselines across multiple medical reasoning benchmarks \cite{kim2025small}. Similarly, Xue et al. \cite{le2025instruction} proposed a QLoRA-based fine-tuning framework for LLaMA-3.1 and Qwen2.5 using CoT-augmented PubMedQA examples, reporting significant improvements in both accuracy and F1 score compared with conventional answer-only instruction tuning.

A separate line of research investigates CoT exclusively as an inference-time prompting strategy without updating model parameters. Wei et al. \cite{wei2022chain,wang2023selfconsistency} first demonstrated that few-shot CoT prompting enables models to generate multi-step reasoning, while self-consistency decoding further improves reasoning performance by aggregating multiple reasoning paths. These prompting techniques have since become foundational methods for medical QA. Building on this paradigm, Med-PaLM and Med-PaLM 2 combine medical instruction tuning with inference-time CoT prompting and self-consistency decoding to achieve state-of-the-art performance on USMLE-style medical reasoning tasks using a 540B-parameter foundation model \cite{singhal2023large,singhal2025toward}. Similarly, Maharjan et al. \cite{maharjan2024openmedlm} proposed OpenMedLM, an automated prompting framework built on the Yi-34B model that leverages elaborate CoT prompting without any task-specific fine-tuning, achieving 72.6\% accuracy on MedQA. Collectively, these studies demonstrate that inference-time reasoning can substantially improve medical QA performance without modifying model parameters, although they rely on relatively large language models and do not teach smaller models to internalize the reasoning process during training.

Despite these advances, important gaps remain in applying rationale-supervised CoT fine-tuning to compact, deployment-oriented medical language models. Most prior work on SLMs has focused on answer-only instruction tuning, whereas rationale-supervised fine-tuning has received comparatively little attention and has not been systematically compared with answer-only training or inference-time CoT prompting under a common experimental framework. Furthermore, previous studies have not isolated the contribution of joint CoT-and-answer supervision, in which models simultaneously learn intermediate clinical reasoning and final answer generation within a unified training objective. 

To the best of our knowledge, this study is among the first to systematically compare joint CoT-and-answer fine-tuning, answer-only fine-tuning, and inference-time CoT prompting without fine-tuning as baselines in medical QA, aiming to identify an effective paradigm for accurate and clinically reasoned responses.

\begin{figure*}[t!]
\centerline{\includegraphics[width=0.9\linewidth]{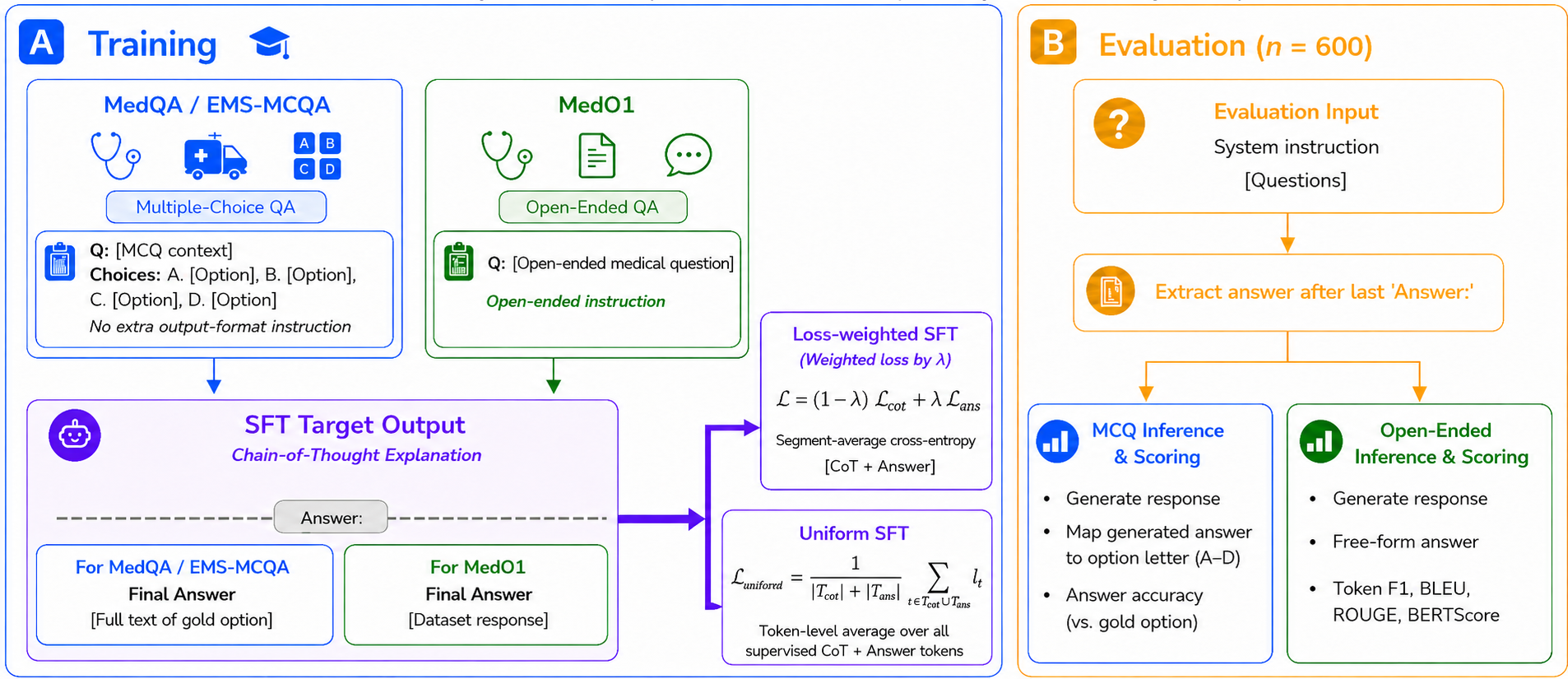}}
\caption{Overview of the training and evaluation pipeline. (A) Training: MCQ (MedQA/EMS-MCQA) and free-text (Med-O1) examples are converted into a shared CoT-and-answer SFT target, supervised under either Uniform (UaS) or Segment-weighted (SwS) loss. (B) Evaluation: generated responses are parsed to output CoT and answer via the \textit{Answer:} delimiter and evaluated by accuracy (MCQ) or Token F1/BLEU/ROUGE/BERTScore (free-text).}
\label{fig:system}
\end{figure*}

\section{Methodology}

Figure \ref{fig:system} presents the proposed method for fine-tuning SLMs (3B–9B) on medical QA using joint CoT-and-answer fine-tuning. Each training example is split into a CoT reasoning trace and a final answer span, with loss computed on both segments. Two strategies are compared: a normalized baseline that applies uniform loss over all tokens and a loss-weighted approach that explicitly balances the reasoning and answer supervision. Both approaches are evaluated across three medical-domain and three general-purpose SLMs to examine how the optimal balance of supervision varies across models.

\subsection{Segmented Supervised Fine-Tuning Targets}

As illustrated in the SFT Target Output component of Figure~\ref{fig:system}, each supervised target sequence is divided into two segments by the literal \texttt{Answer:} marker. Tokens preceding the \texttt{Answer:} marker constitute the CoT explanation trace,
($\mathcal{T}_{\text{cot}}$), which contains the step-by-step clinical reasoning. Tokens following the \texttt{Answer:} marker constitute the final-answer segment,
($\mathcal{T}_{\text{ans}}$), which represents the answer as a sequence of text tokens rather than a single class label. For example, for answer $\mathcal{T}_{\text{ans}}$ \textit{``A. Normal newborn''}, question and multiple-choice option tokens are treated as conditioning context and are excluded from the training loss. Consequently, only tokens in
($\mathcal{T}_{\text{cot}}$) and
($\mathcal{T}_{\text{ans}}$) contribute to parameter updates.

For each supervised output token ($y_t$), let $-\log p_\theta(y_t\mid x, y_{<t})$ denote the token-level cross-entropy loss ($\ell_t$), where ($x$) is the input prompt and ($y_{<t}$) denotes the previously generated target tokens.

For conventional answer-only supervised fine-tuning, the objective is computed exclusively over the answer segment, shown as Eq. \ref{eq:ans_loss}. This objective trains the model to generate the tokenized answer sequence while providing no explicit supervision for the intermediate reasoning process.

\begin{equation}
\label{eq:ans_loss}
\frac{1}{|\mathcal{T}_{\text{ans}}|}
\sum_{t \in \mathcal{T}_{\text{ans}}} \ell_t.
\end{equation}

\begin{table*}[t!]
\caption{Datasets and Statistics}
\label{tab:dataset-overview}
\centering
 \resizebox{1\textwidth}{!}{
\begin{tabular}{lccccc}
\hline
\textbf{Dataset} & \textbf{Answer Type} & \textbf{Ave. CoT Length (\# of Words)} & \textbf{Ave. Answer Length (\# of Words)}&\textbf{Training} &\textbf{Test} \\
\hline

MedQA  & Multiple Choice & 311.5 & 3.5 & 7,080 & 600 \\
EMS-MCQA & Multiple Choice & 51.2 & 5.7 & 2,600 & 600 \\
Med-O1 & Free Text & 312.5 & 94.4 & 13,793 & 600 \\

\hline
\end{tabular}
}
\end{table*}

\subsection{Joint Supervision Objectives}

We investigate two training objectives for jointly supervising the CoT explanation
trace and the final answer span.
Each strategy defines a different weighting scheme over
$\mathcal{T}_{\text{cot}}$ and $\mathcal{T}_{\text{ans}}$, leading to
qualitatively different learning dynamics, particularly when the two segments
differ substantially in length. 

\paragraph{Uniform Token Averaging SFT (UaS)}
The first strategy averages cross-entropy uniformly over all supervised output
tokens $\ell_t$, treating the CoT trace and the final answer as a single undifferentiated
sequence. The calculation is shown in Eq. \ref{eq:standard_loss}, where $\mathcal{T}_{\text{cot}}$ and $\mathcal{T}_{\text{ans}}$ the length of tokens in the CoT and Answer segments, respectively.

\begin{equation}
    \mathcal{L}_{\text{uniform}}
        = \frac{1}{|\mathcal{T}_{\text{cot}}| + |\mathcal{T}_{\text{ans}}|}
          \sum_{t \in \mathcal{T}_{\text{cot}} \cup \mathcal{T}_{\text{ans}}} \ell_t.
    \label{eq:standard_loss}
\end{equation}

\noindent
This formulation provides a natural joint supervision signal: by training on
the full output sequence, the model is exposed to both the intermediate
reasoning steps and the final answer in a single pass.
However, because $|\mathcal{T}_{\text{cot}}| \gg |\mathcal{T}_{\text{ans}}|$
on long-CoT medical data, $\mathcal{L}_{\text{uniform}}$ is numerically
dominated by CoT tokens.
As a result, the gradient pressure on the final answer span is proportionally
small relative to its importance in evaluation, since MedQA accuracy is
determined solely by the predicted answer letter rather than the quality of
the reasoning trace.
This length imbalance motivates the need for an explicit mechanism to
rebalance supervision between the two segments.

\paragraph{Segment-Weighted SFT (SwS)} The second strategy addresses the imbalance in segment lengths by combining the
average losses calculated separately for the two segments.
Let $\ell_t$ denote the cross-entropy loss for output token~$t$.
We compute the segment-average losses for CoT and Answer separately as Eq. \ref{eq:segment_loss}.

\begin{equation}
    \mathcal{L}_{\text{cot}}
        = \frac{1}{|\mathcal{T}_{\text{cot}}|}
          \sum_{t \in \mathcal{T}_{\text{cot}}} \ell_t,
    \qquad
    \mathcal{L}_{\text{ans}}
        = \frac{1}{|\mathcal{T}_{\text{ans}}|}
          \sum_{t \in \mathcal{T}_{\text{ans}}} \ell_t.
    \label{eq:segment_loss}
\end{equation}

%\noindent
%These averages summarize how well the model fits the reasoning trace and the
%final answer span independently.

Then, we calculate the weighted loss using a
tunable answer weight $\lambda \in (0,1]$ as defined in Eq. \ref{eq:weighted_loss}:

\begin{equation}
    \mathcal{L}_{\text{total}}
        = (1 - \lambda)\,\mathcal{L}_{\text{cot}}
        + \lambda\,\mathcal{L}_{\text{ans}},
    \label{eq:weighted_loss}
\end{equation}

\noindent
Because each term is already normalized by its respective segment length,
$\lambda$ directly controls the relative gradient contribution of the two
segments, independently of how long or short each segment happens to be.
The $\lambda$ weight induces a continuous family of training objectives.
At one extreme, when $\lambda = 0$, the training concentrates supervision almost
entirely on the CoT trace, encouraging the model to produce faithful
step-by-step reasoning while receiving minimal direct signal on the answer
token itself.
At $\lambda = 0.5$, both segments contribute equally to the gradient,
providing balanced joint supervision.
At the other extreme, $\lambda = 1.0$ zeroes out the CoT loss entirely: CoT
tokens remain in the target sequence and are still generated during training,
but they receive no gradient, making this setting equivalent to answer-only
supervision.
Any $\lambda \in (0, 1)$ therefore retains some degree of CoT supervision
while simultaneously ensuring that the answer span receives explicit,
length-normalized gradient pressure --- a property that
$\mathcal{L}_{\text{uniform}}$ cannot provide when segment lengths are
highly asymmetric. This tunable-weight formulation follows a broader line of work on balancing multiple loss terms during joint training \cite{kendall2018multi}, adapted here to the specific setting of CoT-and-answer supervision in clinical QA.

\subsection{Inference and Evaluation Protocol}

\paragraph{Inference}
All SLMs share a unified inference procedure regardless of model family,
parameter scale, or domain pretraining background.
We use deterministic generation (greedy decoding) with a maximum output length
of 2048 tokens to avoid prematurely truncating long CoT outputs.
Each model applies the same prompt template, consistent with the
template used during fine-tuning, ensuring that the input format seen at
inference matches the format the adapter was trained on.
The shared \texttt{Answer:} delimiter is used for answer extraction uniformly
across all models and datasets, decoupling the extraction logic from any
model-specific output conventions.

\paragraph{Generalization across SLMs}
The protocol is designed to be model-independent and applies without modification
to all SLMs evaluated in this work, spanning medical-domain models
(BioMistral-7B \cite{labrak2024biomistral}, Meerkat-8B \cite{kim2025small}, Lingshu-7B \cite{xu2025lingshu}) and general-purpose models with no
medical pretraining (Llama-3.2-3B \cite{grattafiori2024llama}, Qwen3.5-4B \cite{qwen35}, Qwen3.5-9B \cite{qwen35}).
The only part that varies across models is the chat template. The same template is used during training and inference, while the loss calculation and final-answer extraction remain unchanged. 
This design ensures that the weighted loss objective in
Eq.~(\ref{eq:weighted_loss}) and the answer extraction procedure transfer
directly to any instruction-tunable SLM that supports LoRA fine-tuning and
a chat template, making the approach readily extensible to future medical-domain or
general-purpose models without protocol modifications.

\paragraph{Answer extraction}
Let $\hat{y}$ denote the generated output. We find the \texttt{Answer:} or \texttt{Final answer:} marker and extract the text after it as the predicted answer $\hat{a}$. For multiple-choice tasks (MedQA, EMS-MCQA), we map $\hat{a}$ to the corresponding option letter $\hat{\ell}$ and compare it with the gold \texttt{answer letter} for evaluation. For open-ended tasks (Med-O1), we retain the full answer $\hat{a}$ for text-based evaluation. The same extraction procedure is used for all SLMs.

\section{Experimental Setup}

\subsection{Datasets}

We evaluate our approach on three public medical QA datasets, each providing
a medical question, a CoT explanation, and a final answer, as
summarized in Table~\ref{tab:dataset-overview}.
The datasets span two answer formats --- multiple choice and free text ---
allowing us to assess the effect of joint CoT-and-answer supervision across
qualitatively different output structures.

\textbf{MedQA}~\cite{jin2021medqa} is a large-scale multiple-choice benchmark
derived from United States Medical Licensing Examination (USMLE) practice
materials.
Each question is paired with four answer options (A--D), a step-by-step CoT
explanation, and a gold answer letter.
We use 7,080 examples for training and 600 for evaluation.
The multiple-choice format makes MedQA well-suited for measuring the direct
impact of $\lambda$ on answer accuracy, since the evaluation signal is precise
and unambiguous.

\textbf{EMS-MCQA}~\cite{ge2026expert} is a multiple-choice dataset grounded in emergency medical
services (EMS) protocols and prehospital care scenarios.
Each question tests clinical decision-making under the constraints of
field-based emergency care, covering topics such as airway management, trauma
triage, and medication administration. This dataset has a relatively short CoT. 

\textbf{Med-O1}~\cite{chen2024huatuogpt} is a free-text medical QA dataset that pairs
clinical questions with long-form CoT reasoning traces and multi-sentence
final answers.
With 13,793 training examples and 600 test examples, it is the largest
dataset in our evaluation and presents a more open-ended generation challenge
than multiple-choice benchmarks. This dataset has the longest CoT and answer in terms of number of words.

\subsection{Hyperparameters}

For each base model and dataset combination, we train one LoRA
adapter~\cite{hu2022lora} per $\lambda \in \{0.2, 0.4, 0.6, 0.8, 1.0\}$ on
the 70\% training split, yielding an independent adapter for every
model--dataset--$\lambda$ triplet with no shared training across datasets or
$\lambda$ values.
We minimize Eq.~(\ref{eq:weighted_loss}) using AdamW \cite{loshchilov2017decoupled} with a learning rate of
$2 \times 10^{-5}$, cosine decay with 3\% linear warmup, batch size~1,
gradient accumulation over 16 steps (effective batch size~16), maximum
sequence length~2048, and 3 training epochs.
LoRA adapters are configured with rank~16, alpha~16, and dropout~0.05.
All experiments use bf16 precision with gradient checkpointing on H200 GPUs.
Only LoRA parameters are updated during training; the frozen backbone
is shared between the zero-shot baseline and all fine-tuned evaluations,
ensuring that any performance difference is attributable solely to the
adapter and the choice of $\lambda$.

\subsection{Evaluation}

For each base model and dataset, we compare three conditions:

\begin{itemize}
    \item \textbf{Baseline without SFT (Base).} The frozen base model is evaluated
    without a LoRA adapter to assess the existing medical knowledge of the model and ability to follow instructions prior to any
    task-specific fine-tuning.

    \item \textbf{Uniform Token Averaging SFT (UaS).} A single LoRA adapter is trained using
    the normalized weighted objective in Eq.~(\ref{eq:standard_loss}), which
    applies equal token-level weighting across the full CoT-and-answer output
    sequence. This condition serves as a reference to isolate the effect of
    length-normalization from the effect of explicit segment reweighting.

    \item \textbf{Segment-Weighted SFT (SwS).} LoRA adapters are trained for each model
    using the weighted objective in Eq.~(\ref{eq:weighted_loss}), one per
    $\lambda \in \{0.2, 0.4, 0.6, 0.8, 1.0\}$, with the \texttt{Answer:}
    delimiter used to separate $\mathcal{T}_{\text{cot}}$ from
    $\mathcal{T}_{\text{ans}}$ at training time. This condition directly
    measures the effect of varying the supervision balance between the CoT
    trace and the final answer span. 
\end{itemize}

\noindent

For MedQA and EMS-MCQA, the primary metric is the accuracy based on the letter answer.
For Med-O1, we report token-F1, BLEU-2, ROUGE-L, and BERTScore F1 to capture both
lexical overlap and semantic similarity of the generated free-text responses.

\section{Results}
\label{sec:results}

\subsection{Performance Comparison on MCQ datasets}

Table \ref{tab:comparisonMCQ}  compares general-purpose and medical models on MedQA and EMS-MCQA under the Base, UaS, and SwS settings. For each model, we compared its best-performing setting against the corresponding baseline using a paired Wilcoxon signed-rank test on the test dataset; results marked with * indicate a statistically significant improvement (p $<$ 0.01). 

On MedQA, the largest gains come from the general-purpose Qwen models: Qwen3.5-4B rises from a 33.0\% baseline to 65.5\% with UaS and 77.8\% with SwS (a 44.8\% gain), while the performance of Qwen3.5-9B increases from 35.5 to 74.5. Meerkat-8B posts the highest MedQA baseline at 80.5\%, likely because it was fine-tuned directly on USMLE-style medical QA data and already reasons effectively without additional finetuning. And its score actually drops to 75.3\% under UaS and only partially recovers to 76.0\% under SwS, suggesting that the models have less headroom to benefit from the proposed finetuning strategies. Lingshu-7B changes little, with UaS setting (66.3\%) marginally ahead of SwS setting (65.5\%), whereas BioMistral-7B improves substantially, from 30.2\% to 47.2\%. Overall, Qwen3.5-4B achieves the top score on MedQA at 77.8\% ($\lambda$=0.8).

A similar pattern is observed on EMS-MCQA, where the SwS setting improves the performance of every model over its baseline, with the largest gains observed for models with lower baseline performance. Qwen3.5-4B jumps from 33.2\% to 62.5\%, and Qwen3.5-9B from 32.7\% to 57.3\%. Among the medical models, Lingshu-7B leads with 73.8\% ($\lambda$=0.8), just ahead of Meerkat-8B at 73.2\%, and BioMistral-7B improves from 50.3\% to 58.8\%. Llama-3.2-3B is the one exception with a dip under UaS setting (from 66.0\% to 62.0\%), though SwS recovers and slightly exceeds baseline at 66.2\%. Notably, since $\lambda$=1 corresponds to fine-tuning on answer-only supervision, the fact that both Llama-3.2-3B and Meerkat-8B achieve their best EMS-MCQA results at $\lambda$=1.0 indicates that joint CoT-and-answer supervision offers little to no additional benefit for these two models on this dataset.

These results show SwS is especially effective at boosting general-purpose models with weaker baselines, letting them approach — and on MedQA, sometimes surpass — medical-domain models.

\begin{table*}[h!]
  \caption{Performance Comparison on MCQ Datasets: MedQA and EMS-MCQA}
  \label{tab:comparisonMCQ}
  \centering
  \resizebox{0.55\textwidth}{!}{%
    \begin{tabular}{l|ccc|ccc}
      \hline
      \multirow{2}{*}{\textbf{Models}} & \multicolumn{3}{c|}{\textbf{MedQA}} &
      \multicolumn{3}{c}{\textbf{EMS-MCQA}} \\\cline{2-7}
      & Base & UaS & SwS ($\lambda$) & Base & UaS & SwS ($\lambda$) \\
      \hline
      \multicolumn{7}{l}{\textbf{General-purpose Models}} \\\hline
      Llama-3.2-3B  & 58.0 & 63.5 & \textbf{64.7*} (0.6) & 66.0 & 62.0 & \textbf{66.2} (1.0) \\
      Qwen3.5-4B    & 33.0 & 65.5 & \textbf{77.8*} (0.8) & 33.2 & 53.8 & \textbf{62.5*} (0.8) \\
      Qwen3.5-9B    & 35.5 & 74.2 & \textbf{74.5*} (0.2) & 32.7 & 48.2 & \textbf{57.3*} (0.8) \\
      \hline
      \multicolumn{7}{l}{\textbf{Medical-domain Models}} \\\hline
      BioMistral-7B & 30.2 & 44.8 & \textbf{47.2*} (0.2) & 50.3 & 58.2 & \textbf{58.8*} (0.6) \\
      %Hulu-Med-4B   & 16.3 & 69.0 & 70.0 (0.8) & 35.0 & --$^\ddagger$ & 28.7 (1.0) \\
      Lingshu-7B    & 64.2 & \textbf{66.3} & 65.5 (1.0) & 67.3 & 72.8 & \textbf{73.8*} (0.8) \\
      %II-Medical-8B & 92.2 & 77.3 & 93.3 (1.0) & 82.8 & 78.8 & 82.8 (1.0) \\
      %Meerkat-7B    & 82.7 & 76.8 & 78.0 (0.2) & 69.3 & 67.5 & 67.7 (0.8) \\
      Meerkat-8B    & \textbf{80.5} & 75.3 & 76.0 (0.4) & 70.2 & 71.3 & \textbf{73.2*} (1.0) \\
      \hline
    \end{tabular}%
  }
\end{table*}

\begin{table*}[h!]
\caption{Performance Comparison on Med-O1 Dataset}
\label{tab:medo1_results}
\centering
\resizebox{\textwidth}{!}{%
\begin{tabular}{l|cccccccccccc}
\hline
 & \multicolumn{3}{c}{Token F1} & \multicolumn{3}{c}{BLEU-2} &  \multicolumn{3}{c}{ROUGE-L} & \multicolumn{3}{c}{BERTScore F1} \\
\cline{2-13}
\textbf{Models} & Base & UaS & SwS ($\lambda$) & Base & UaS & SwS ($\lambda$) &  Base & UaS & SwS ($\lambda$) & Base & UaS & SwS ($\lambda$) \\
\hline
\multicolumn{13}{l}{\textbf{General-purpose models}} \\\hline
Llama-3.2-3B & 39.6 & \textbf{48.7*} & 47.6 (0.4) & 12.7 & \textbf{28.1*} & 27.1 (0.2) & 26.4 & \textbf{34.4*} & 33.9 (0.2) & 87.9 & \textbf{89.5*} & 89.4 (0.4) \\
Qwen3.5-4B & 24.6 & \textbf{29.1*} & 28.0 (0.2) & 6.3 & \textbf{10.5*} & 9.7 (0.2) & 12.9 & \textbf{16.8*} & 16.3 (0.2) & 83.7 & \textbf{85.1*} & 85.0 (0.2) \\
Qwen3.5-9B & 26.3 & 29.7  & \textbf{30.4*} (1.0) & 8.2 & \textbf{11.4*} & \textbf{11.4*} (0.4) & 14.9 & 17.5 & \textbf{18.5*}  (1.0) & 84.2 & \textbf{85.3*}  & \textbf{85.3*} (0.4) \\
\hline
\multicolumn{13}{l}{\textbf{Medical-domain models}} \\\hline
BioMistral-7B & 33.5 & \textbf{47.6*} & 44.8 (0.4) & 11.2 & \textbf{26.7*} & 24.1 (0.4) & 23.0 & \textbf{33.8*} & 31.7 (0.4) & 86.9 & \textbf{89.3*} & 89.0 (0.4) \\
%Hulu-Med-4B & 26.5 & \textbf{53.0} & 51.4 (0.8) & 22.9 & \textbf{52.8} & 51.1 (0.6) &  18.1 & \textbf{38.8} & 37.8 (0.6) & 87.3 & \textbf{90.4} &  90.3 (0.8)\\
Lingshu-7B & 30.1 & \textbf{48.6*} &  46.6 (0.6) & 7.9 & \textbf{29.0*} & 26.9 (0.8) & 21.1 & \textbf{34.9*} & 33.7 (0.6) & 86.8 & \textbf{89.6*} & 89.4 (0.6) \\
%II-Medical-8B & 9.1 & \textbf{53.3} & 52.9 (0.4) & 7.0 & \textbf{53.3} & 53.1 (0.4) & 6.3 & \textbf{38.3} & 38.0 (0.4) & 82.1 & 90.4 & \textbf{90.5} (0.8) \\
%Meerkat-7B & 22.5 & 44.5 & \textbf{47.2} (1.0) & 19.2 & 41.1 & \textbf{45.2} (1.0) & 15.7 & 28.8 & \textbf{30.6} (1.0) & 85.6 & 89.4 & \textbf{89.5} (1.0) \\
Meerkat-8B & 28.1 & \textbf{51.5*} & 51.4 (0.8) & 7.1 & \textbf{30.8*} & 30.7 (0.4) & 19.2 & \textbf{37.0*} & 36.7 (0.4) & 86.3 & \textbf{90.0*} & \textbf{90.0*} (0.2) \\
\hline
\end{tabular}%
}
\end{table*}

\subsection{Performance Comparison on Free-Text Answer dataset}

Table \ref{tab:medo1_results} presents the performance of general-purpose and medical-domain models on the Med-O1 free-text answer dataset, evaluated with Token F1, BLEU-2, ROUGE-L, and BERTScore F1. Across nearly all models and metrics, both UaS and SwS improve substantially over the baseline. For each model, we compared its best-performing setting against the corresponding baseline using a paired Wilcoxon signed-rank test on the test dataset; results marked with * indicate a statistically significant improvement (p $<$ 0.01). 

Among the general-purpose models, Llama-3.2-3B achieves the strongest overall performance. Under UaS, its Token F1 rises from 39.6 to 48.7, BLEU-2 from 12.7 to 28.1, ROUGE-L from 26.4 to 34.4, and BERTScore F1 from 87.9 to 89.5 — outperforming both Qwen3.5-4B and Qwen3.5-9B, which improve across all metrics but remain below the performance of Llama-3.2-3B.

The medical-domain models show particularly large gains over baseline. Meerkat-8B delivers the best results in the table overall, reaching 51.5 Token F1, 30.8 BLEU-2, and 37.0 ROUGE-L under UaS, and the highest BERTScore F1 of 90.0 under both UaS and SwS ($\lambda$=0.2). Relative to its baseline, Meerkat-8B improves by 23.4 points in Token F1, 23.7 in BLEU-2, and 17.8 in ROUGE-L — the largest gains of any model in the table — underscoring the effectiveness of the proposed methods for medical free-text answer generation. Lingshu-7B and BioMistral-7B follow a similar pattern, with UaS consistently outperforming SwS across all four metrics for both models.

Overall, both UaS and SwS improve answer quality across general-purpose and medical-domain models, with UaS generally producing the strongest gains, especially for the medical-domain models.

%\subsection{Performances on Questions in Sub-Domains}

% \begin{table}[t]
% \caption{Performance Comparison on Medical Sub-Domains}
% \label{tab:category-delta}
% \centering

% \begin{tabular}{lcccc}
% \hline
% Category & \# of Q. & Base & UaS & SwS\\
% \hline
% Neurology and Psychiatry & 83 && \\
% Endocrinology and Metabolism & 67 && \\
% Pediatrics & 65 && \\
% Cardiovascular Disorders & 60 && \\
% Infectious Diseases & 58 && \\
% Hematology and Oncology & 54 && \\
% Obstetrics and Gynecology & 52 && \\
% Gastroenterology and Hepatology & 44 && \\
% Musculoskeletal and Rheumatology & 25 && \\
% Surgery and Trauma & 24 && \\
% Pulmonology & 22 && \\
% Other& 46 && \\
% \hline
% \end{tabular}
% \end{table}

\begin{table*}[t]
\caption{MedQA sub-domain accuracy for Qwen3.5-4B and Meerkat-8B.}
\label{tab:category-delta}
\centering
\setlength{\tabcolsep}{5pt}
\small
\begin{tabular}{l c ccc ccc}
\hline
Category & \#Q
& \multicolumn{3}{c}{Qwen3.5-4B}
& \multicolumn{3}{c}{Meerkat-8B} \\
\cmidrule(lr){3-5}\cmidrule(lr){6-8}
& & Base & UaS & SwS (0.8)
  & Base & UaS & SwS (0.4) \\
\hline
Neurology / Psychiatry      & 83 & 33.7 & 61.4 & \textbf{80.7} & 76.4 & 76.4 & 76.4 \\
Endocrinology / Metabolism & 67 & 23.9 & 65.7 & 82.1 & \textbf{83.1} & 81.4 & 81.4 \\
Pediatrics                  & 65 & 36.9 & 70.8 & 78.5 & \textbf{85.2} & 78.7 & 82.0 \\
Cardiovascular              & 60 & 31.7 & 68.3 & 73.3 & \textbf{75.9} & 61.1 & 63.0 \\
Infectious Diseases         & 58 & 37.9 & 67.2 & \textbf{82.8} & 75.6 & 71.1 & 71.1 \\
Hematology / Oncology      & 54 & 31.5 & 75.9 & 81.5 & \textbf{82.4} & 80.4 & 78.4 \\
OB/GYN                      & 52 & 32.7 & 57.7 & 69.2 & 73.7 & \textbf{84.2} & 78.9 \\
Gastroenterology            & 44 & 25.0 & 59.1 & 70.5 & \textbf{80.5} & 65.9 & 63.4 \\
MSK / Rheumatology         & 25 & 32.0 & 76.0 & 76.0 & \textbf{88.9} & \textbf{88.9} & 83.3 \\
Surgery / Trauma            & 24 & 37.5 & 58.3 & \textbf{83.3} & 75.0 & 60.0 & 70.0 \\
Pulmonology                 & 22 & 40.9 & 50.0 & 77.3 & 80.0 & 75.0 & \textbf{90.0} \\
Other                       & 46 & 39.1 & 67.4 & 76.1 & \textbf{81.6} & 77.6 & 79.6 \\
\hline
\end{tabular}
\end{table*}
\begin{figure*}[h!]
\centering
\includegraphics[width=\textwidth]{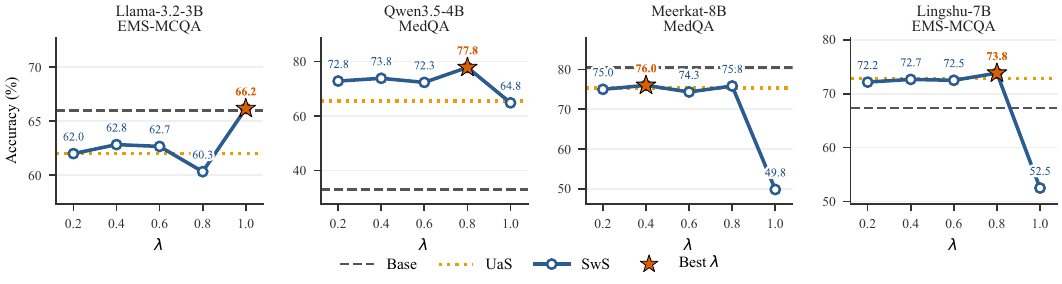}
\caption{Effect of $\lambda$ on SwS performance for the top-performing models on MedQA and EMS-MCQA.}
\label{fig:lambda-ablation}
\end{figure*}

\subsection{Performances on Questions in Clinical Domains}
To examine clinical specialty-level performance, we used the GPT-5.2 model to categorize the 600 questions in the MedQA test set into 12 medical sub-domains based on their primary clinical topic. Qwen3.5-4B and Meerkat-8B were selected because they are the best-performing general-purpose and medical-domain models, respectively, in the overall MedQA evaluation. The bold values in Table \ref{tab:category-delta} indicate the highest accuracy across the two models and their three evaluation settings within each sub-domain.

Although Qwen3.5-4B performs worse than Meerkat-8B under the baseline setting, SwS closes this gap in several specialties. In particular, Qwen3.5-4B with SwS achieves the best accuracy in Neurology/Psychiatry (80.7\%), Infectious Diseases (82.8\%), and Surgery/Trauma (83.3\%), exceeding the best Meerkat-8B result in these categories by 4.3\%, 7.2\%, and 8.3\%, respectively. It also approaches the best Meerkat-8B performance in Endocrinology/Metabolism, Cardiovascular, and Hematology/Oncology. Meerkat-8B, however, remains stronger in categories such as Pediatrics, OB/GYN, Gastroenterology, MSK/Rheumatology, and Pulmonology.
A possible explanation is that SwS helps compensate for the limited medical specialization of the general-purpose model by eliciting relevant knowledge and promoting more systematic clinical reasoning. This support may be particularly effective in sub-domains whose questions contain clear diagnostic patterns or require multi-step reasoning. 

Although the base Meerkat-8B model achieves the highest overall performance among the settings tested, some sub-domains still benefit from the proposed strategies. Pulmonology accuracy improves from 80.0\% at baseline to 90.0\% under SwS ($\lambda$=0.4), indicating that joint CoT-and-answer fine-tuning can further strengthen the performance of Meerkat-8B even in a sub-domain where it is already the strongest model. OB/GYN accuracy also improves substantially, from 73.7\% at baseline to 84.2\%, though this gain comes under UaS rather than SwS.

\section{Ablation Studies}

\subsection{Effect of the Segment Weight $\lambda$}

Figure \ref{fig:lambda-ablation} shows an ablation study on the parameter $\lambda$ under the SwS setting, for the best-performing general-purpose and medical-domain model on each of the two datasets: Llama-3.2-3B and Lingshu-7B on EMS-MCQA, and Qwen3.5-4B and Meerkat-8B on MedQA. For each model, the dashed line marks the Base accuracy, the dotted line marks the UaS accuracy, and the solid curve traces SwS accuracy as $\lambda$ is varied from 0.2 to 1.0, with the star indicating the best-performing $\lambda$.

The four models exhibit distinct sensitivity patterns to $\lambda$. Llama-3.2-3B on EMS-MCQA improves only modestly for $\lambda$ between 0.2 and 0.6 (62.0\% to 62.7\%), dips slightly at $\lambda$=0.8 (60.3\%), then rises sharply at $\lambda$=1.0 to its best result of 66.2\% — the only model among the four whose optimal $\lambda$ is 1.0, corresponding to answer-only fine-tuning. Qwen3.5-4B on MedQA shows a clearer peak, climbing from 72.8\% at $\lambda$=0.2 to a maximum of 77.8\% at $\lambda$=0.8, before dropping to 64.8\% at $\lambda$=1.0. Meerkat-8B on MedQA stays relatively flat and close to its UaS level across $\lambda$=0.2–0.6 (74.3\%–76.0\%), reaches its best result of 76.0\% at $\lambda$=0.4, and then falls sharply to 49.8\% at $\lambda$=1.0 — well below its base accuracy. Lingshu-7B on EMS-MCQA follows a similar pattern to Qwen3.5-4B, with accuracy gradually increasing from 72.2\% at $\lambda$=0.2 to a peak of 73.8\% at $\lambda$=0.8.

Across all four models, the results indicate that a moderate blend of CoT and answer supervision ($\lambda$=0.4–0.8) generally yields the best or near-best performance, while relying exclusively on answer-only supervision ($\lambda$=1.0) is unstable, and sometimes falls below even the base model accuracy.

\subsection{Effect of the Length of Chain-of-Thought}

Since our approach does not explicitly constrain the output CoT length, we further examine whether performance differs when the generated CoT is shorter than, comparable to, or longer than the reference CoT - a question motivated by growing evidence that longer reasoning chains do not monotonically improve accuracy and can instead degrade performance once reasoning length exceeds a model- and task-specific threshold \cite{wu2026more,su2025between}. For each model–dataset pair, we retain its best-performing setting and group examples by the ratio of generated-to-reference CoT length into Short (S) ($<$0.5×), Medium (N) (0.5–2×), and Long (L) ($>$2×) bins.

\begin{figure*}[t]
\begin{minipage}{0.245\textwidth}
\centering
\includegraphics[width=\linewidth]{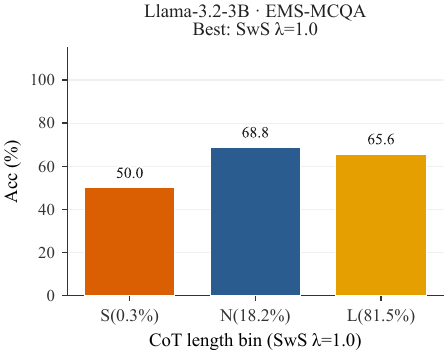}
\end{minipage}\hfill
\begin{minipage}{0.245\textwidth}
\centering
\includegraphics[width=\linewidth]{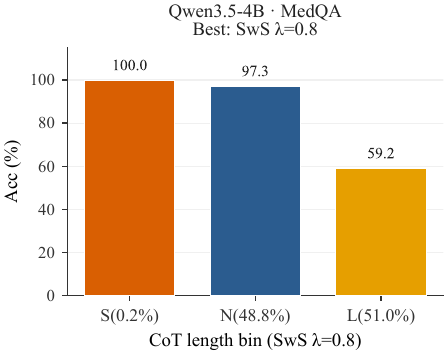}
\end{minipage}\hfill
\begin{minipage}{0.245\textwidth}
\centering
\includegraphics[width=\linewidth]{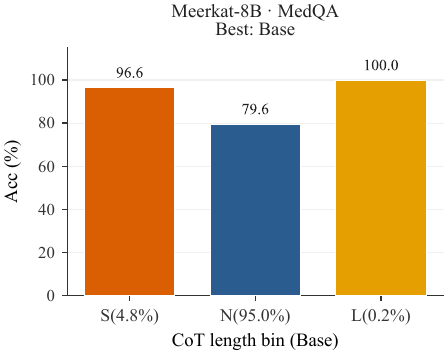}
\end{minipage}\hfill
\begin{minipage}{0.245\textwidth}
\centering
\includegraphics[width=\linewidth]{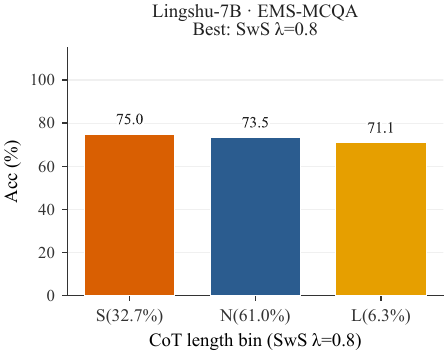}
\end{minipage}
\caption{MCQ accuracy by generated-to-reference CoT length bin, shown under each model's best-performing setting: Llama-3.2-3B/EMS-MCQA (SwS, $\lambda$=1.0), Qwen3.5-4B/MedQA (SwS, $\lambda$=0.8), Meerkat-8B/MedQA (Base), and Lingshu-7B/EMS-MCQA (SwS, $\lambda$=0.8). Percentages in parentheses on the x-axis labels indicate the share of the test dataset falling into each bin.}
\label{fig:lenbin-mcq}
\end{figure*}

On MCQ datasets, the relationship between CoT length and accuracy varies across settings. Qwen3.5-4B on the MedQA dataset exhibits the clearest difference: Medium-length outputs achieve 97.3\% accuracy, compared with 59.2\% for Long outputs, even though the two bins contain similar proportions of the test set (48.8\% and 51.0\%, respectively). For Lingshu-7B on EMS-MCQA dataset, accuracy changes more gradually, from 75.0\% for Short length CoT outputs (32.7\% of examples) to 73.5\% for Medium (61.0\%) and 71.1\% for Long (6.3\%). Llama-3.2-3B on EMS-MCQA dataset produces Long CoTs for 81.5\% of examples, with 65.6\% accuracy in this bin compared with 68.8\% for Medium-length outputs (18.2\%) and 50.0\% for the rare Short bin (0.3\%). For Meerkat-8B on the MedQA dataset, 95.0\% of examples fall into the Medium bin, where accuracy is comparatively lower (79.6\%) than in the small Short (4.8\%, 96.6\%) and Long (0.2\%, 100.0\%) bins; given how few examples populate these extremes, we think this pattern likely reflects noise rather than a reliable length effect.

\begin{figure}[t]
\centering
\begin{minipage}{0.48\columnwidth}
\centering
\includegraphics[width=\linewidth]{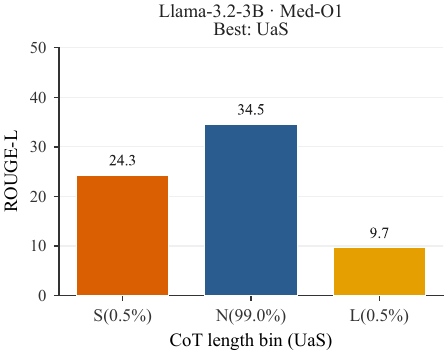}
\end{minipage}\hfill
\begin{minipage}{0.48\columnwidth}
\centering
\includegraphics[width=\linewidth]{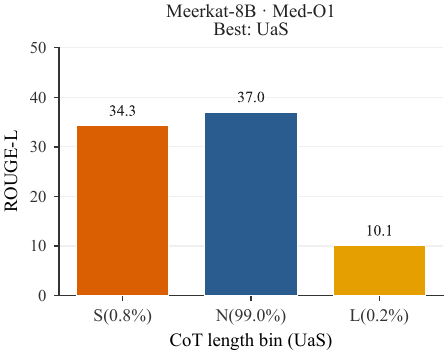}
\end{minipage}
\caption{Med-O1 ROUGE-L by generated-to-reference CoT length bin, shown under each model's best-performing setting (UaS): Llama-3.2-3B (left) and Meerkat-8B (right). Percentages in parentheses on the x-axis labels indicate the share of the test dataset falling into each bin.}
\label{fig:lenbin-medo1}
\end{figure}

On the Med-O1 dataset, 99.0\% of examples fall into the Medium bin for both Llama-3.2-3B and Meerkat-8B under their best UaS setting. Their Medium-bin ROUGE-L scores of 34.5 and 37.0 are correspondingly close to the overall UaS scores of 34.4 and 37.0 reported in Table~\ref{tab:medo1_results}. The Short and Long bins each contain under 1\% of examples for both models, but both models show a similar pattern: ROUGE-L drops sharply in the Long bin (9.7 for Llama-3.2-3B, 10.1 for Meerkat-8B) relative to Medium, while Short-bin scores fall in between (24.3 and 34.3, respectively) — though given how few examples populate these extreme bins, this comparison should be read as suggestive rather than conclusive.

Overall, CoT length is associated with performance in some model–task settings, but the results do not indicate a single optimal reasoning-length across models.

\section{Discussion}
\label{sec:discussion}

\subsection{Uniform vs. Segment-Weighted Supervision: A Task-Dependent Trade-off in Medical QA}

When the SwS setting is applied to the MCQ datasets, $L_{ans}$ is computed during training against one specific, fixed option text drawn from a small, closed set of known alternatives (labeled A through D). Even though the model must generate this text in free form, the target itself is highly constrained and low-entropy. There is only a handful of valid strings exist for the answer segment to converge toward. Upweighting $L_{ans}$ via $\lambda$ therefore provides a clean, low-noise training signal, pushing the model firmly toward one of a few discrete target answers. For the Med-O1 dataset, the answers are open-ended free text with no closed alternative set. Because there is no discrete boundary for the extra supervision to sharpen by choosing a correct one from a set, upweighting $L_{ans}$ does not help the model to generate correct free-text answer. Changing the $L_{ans}$ simply shifts training capacity toward reproducing one particular phrasing of the answer more precisely, at the expense of fluency and coherence in the CoT segment.

When the UaS setting is applied to the MCQ datasets, the $L_{uniform}$ is computed as a single token-level average over the entire CoT-and-answer sequence, with no explicit mechanism distinguishing the answer segment from the reasoning that precedes it. Because CoT explanations are typically far longer than the fixed-length answer segment, the contribution of the answer to the overall loss is diluted in proportion to how long the reasoning chain happens to be. As a result, UaS provides a weaker signal for the decision that matters most at evaluation. For the Med-O1 dataset, this uniform weighting of UaS setting is well matched to the task structure. Because both the CoT and answer are open-ended text with no fixed set of possible answers, there is no separate loss calculation for the model to be pushed toward. The entire sequence is being evaluated which avoids the answer-segment overfitting that SwS could induce.

\subsection{General-Purpose vs. Medical-Domain Pretraining: A Model-Dependent Response to SFT}

On MCQ tasks, general-purpose models show far larger gains, while medical-domain models show diminishing or even negative returns. On Med-O1, however, the pattern shifts: both model families benefit substantially, and the domain gap narrows rather than persists.

Two factors likely explain this difference. First, medical-domain pretraining, as used in models such as Meerkat and Lingshu, may provide stronger prior medical knowledge and familiarity with medical QA. This advantage is particularly relevant for MCQs, where performance depends heavily on identifying the correct option. In contrast, Med-O1 requires the generation of longer, reference aligned explanations, leaving more room for improvement through supervised fine tuning of reasoning and response structure. Second, general-purpose models lack both medical knowledge and response-formatting skills, so they have more room to improve on both tasks. However, their improvement on MCQs is limited by how much of the accuracy gap comes from missing medical knowledge. Training can improve formatting and reasoning more directly, but it can only partially compensate for missing domain knowledge. This is why the gains are larger for Med-O1 than for MCQs.

These results suggest that the proposed fine-tuning approach is better at reducing formatting and reasoning gaps than domain-knowledge gaps. For MCQs, medical knowledge is critical, so medical-domain pretraining is difficult to replace through fine-tuning alone. For free-text generation, however, performance depends more on producing the right structure and style, which the proposed training can improve across different base models. This makes joint CoT-and-answer SFT especially useful for adapting general-purpose models to clinical QA, allowing smaller general models to achieve performance similar to much larger medical-domain models.

\subsection{Enforcing CoT Length: A Model-Dependent Response to SFT}

The results on analyzing the effect of CoT length suggest that excessively long generations are a consistent failure mode across both MCQ and free-text tasks and across different model families. This suggests that controlling CoT length during training could help reduce this problem. However, the optimal length varies across models, so a single fixed length constraint should not be applied to all models. A global constraint could suppress generation patterns that are harmless or even beneficial for some models.

Instead of using a hard length constraint during fine-tuning, we suggest a soft, length-aware penalty (e.g., a penalty that increases when the generated CoT substantially exceeds the reference length) to discourage excessively long reasoning traces. This penalty could be applied selectively to model–task combinations where long outputs are both common and strongly associated with lower accuracy. For example, Qwen3.5-4B on MedQA dataset has more than half of its test outputs in the Long CoT, with nearly a 40-point drop in accuracy. For model–task pairs where long CoT outputs are rare or do not substantially hurt performance (e.g., Meerkat-8B, etc.), length constraints are unlikely to provide meaningful benefits and may unnecessarily restrict useful generation patterns.

Therefore, we view CoT length control as an adjustment based on specific models and tasks rather than a universal training modification. An analysis of the relationship between CoT length and performance can first identify the model and task combinations that may benefit from length regularization, which can then be applied selectively to those cases.

\subsection{Limitations}

This study has several limitations that suggest directions for future work. First, our experiments use a single training seed, LoRA-based fine-tuning, and models up to 9B parameters. Although using a fixed 600-sample test set provides a controlled comparison, some CoT length bins contain very few examples, limiting the statistical power of the analysis. Future work should use larger test sets, multiple random seeds, and paired significance tests to better validate the relationship between CoT length and accuracy. Second, our results show that the effectiveness of UaS and SwS depends on both task type (MCQ vs. free-text) and model domain (general vs. medical). However, we evaluate only six models, two supervision schemes, and two task types. A larger range of models, datasets, and clinical categories is needed to determine whether these findings generalize to other medical QA tasks, such as multi-label classification and structured extraction. Third, excessively long CoT generation is associated with lower performance for some model–task pairs, but this effect varies across models. We do not directly evaluate length-aware training in this study. Future work should test soft length penalties, such as adaptive length-based reward shaping \cite{rakotonirina2026correct} or inference-time reasoning-length control \cite{eisenstadt2025overclocking}, and investigate whether they can be applied selectively based on simple diagnostics. Finally, the SwS weight $\lambda$ is tuned separately for each model–dataset pair using grid search, which is computationally expensive. Future work could explore adaptive or learned $\lambda$ scheduling based on factors such as model domain, task type, or CoT-length distribution to reduce tuning costs while maintaining the benefits of task-specific weighting.

\FloatBarrier
\normalfont\normalsize
\section{Conclusion}

We investigate how to balance CoT and answer supervision when fine-tuning small medical QA models, comparing uniform supervision with segment-weighted supervision. Across six general-purpose and medical-domain models and both MCQ and free-text QA tasks, we find that the best supervision strategy depends on the task and the model. For MCQs, SwS generally works best, especially for general-purpose models, while medical-domain models show smaller or even negative gains. For free-text QA, UaS performs best and provides consistent gains across both model types, suggesting that uniform supervision better fits open-ended answer generation. These results indicate that fine-tuning can more easily reduce gaps in reasoning and response structure than gaps in domain knowledge. We also find that excessively long CoT is linked to lower performance in some model–task settings, suggesting that length-aware training may help in selected cases. Overall, the choice between UaS and SwS should depend on the task structure and the knowledge already captured by the base model, rather than using one strategy for all models and tasks.

%\clearpage
%\raggedbottom
\normalfont\normalsize
\bibliographystyle{IEEEtran}
\bibliography{references}

\end{document}